\documentclass[11pt,a4paper]{article}
\usepackage[hyperref]{acl2020}
\usepackage{times}
\usepackage{latexsym}

\usepackage{booktabs}
\usepackage{amsmath}
\usepackage{multirow}
\usepackage{xspace}
\usepackage[inline]{enumitem}
\usepackage{boxedminipage}
\usepackage{soul}

\usepackage{comment}

\newcommand{\dataset}{\textsc{SciREX}\xspace}

\newcommand{\scierc}{\textsc{SciERC}\xspace}

\newcommand{\dygie}{\textsc{DyGIE++}\xspace}

\usepackage{xcolor}

\definecolor{method}{HTML}{d5a6bd}
\definecolor{task}{HTML}{ffe599}
\definecolor{metric}{HTML}{b6d7a8}
\definecolor{dataset}{HTML}{9fc5e8}

\usepackage{microtype}
\usepackage{graphicx}
\aclfinalcopy % Uncomment this line for the final submission
\title{\dataset: A Challenge Dataset for Document-Level  \\ Information Extraction}

\author{Sarthak Jain$^{2}$\thanks{\enspace Work done while at AI2}  \quad Madeleine van Zuylen$^{1}$ \quad  Hannaneh Hajishirzi$^{1,3}$ \quad Iz Beltagy$^{1}$\\
Allen Institute for AI$^{1}$ \quad Northeastern University$^{2}$ \quad University of Washington$^{3}$ \\
\texttt{jain.sar@northeastern.edu}\\  \texttt{\{madeleinev,hannah,beltagy\}@allenai.org}\\
}

\date{}

\begin{document}
\maketitle
\begin{abstract}
Extracting information from full documents is an important problem in many domains, but most previous work focus on identifying relationships within a sentence or a paragraph. It is challenging to create a large-scale information extraction (IE) dataset at the document level since it requires an understanding of the whole document to annotate entities and their document-level relationships that usually span beyond sentences or even sections. In this paper, we introduce \dataset, a  document level IE dataset that encompasses multiple IE tasks, including salient entity identification and document level $N$-ary relation identification from scientific articles.
We annotate our dataset by integrating automatic and human annotations, leveraging existing scientific knowledge resources. We develop a neural model as a strong baseline that extends previous state-of-the-art IE models to document-level IE. Analyzing the model performance shows a significant gap between human performance and current baselines, inviting the community to use our dataset as a challenge to develop document-level IE models.  Our data and code are  publicly available at \url{https://github.com/allenai/SciREX}
\end{abstract}

\section{Introduction}

Extracting information about entities and their relationships from unstructured text is an important problem in NLP. Conventional datasets and methods for information extraction (IE) focus on within-sentence relations from general Newswire text~\cite{zhang2017tacred}. However, recent work started studying the development of
full IE models and datasets for short paragraphs
(e.g., information extraction from abstracts of scientific articles as in \scierc~\cite{luan2018multitask}), 
or only extracting relations (given ground truth entities)
on long documents (e.g.~\citet{jia2019document}). 
While these tasks provide a reasonable testbed for developing IE models, a significant amount of information can only be gleaned from analyzing the full document.    To  this  end,  not much work has been done on developing full IE datasets and model
for long documents. 

\begin{figure}[t]
    \centering
    \begin{boxedminipage}{\columnwidth}
    \includegraphics[width=.8\columnwidth]{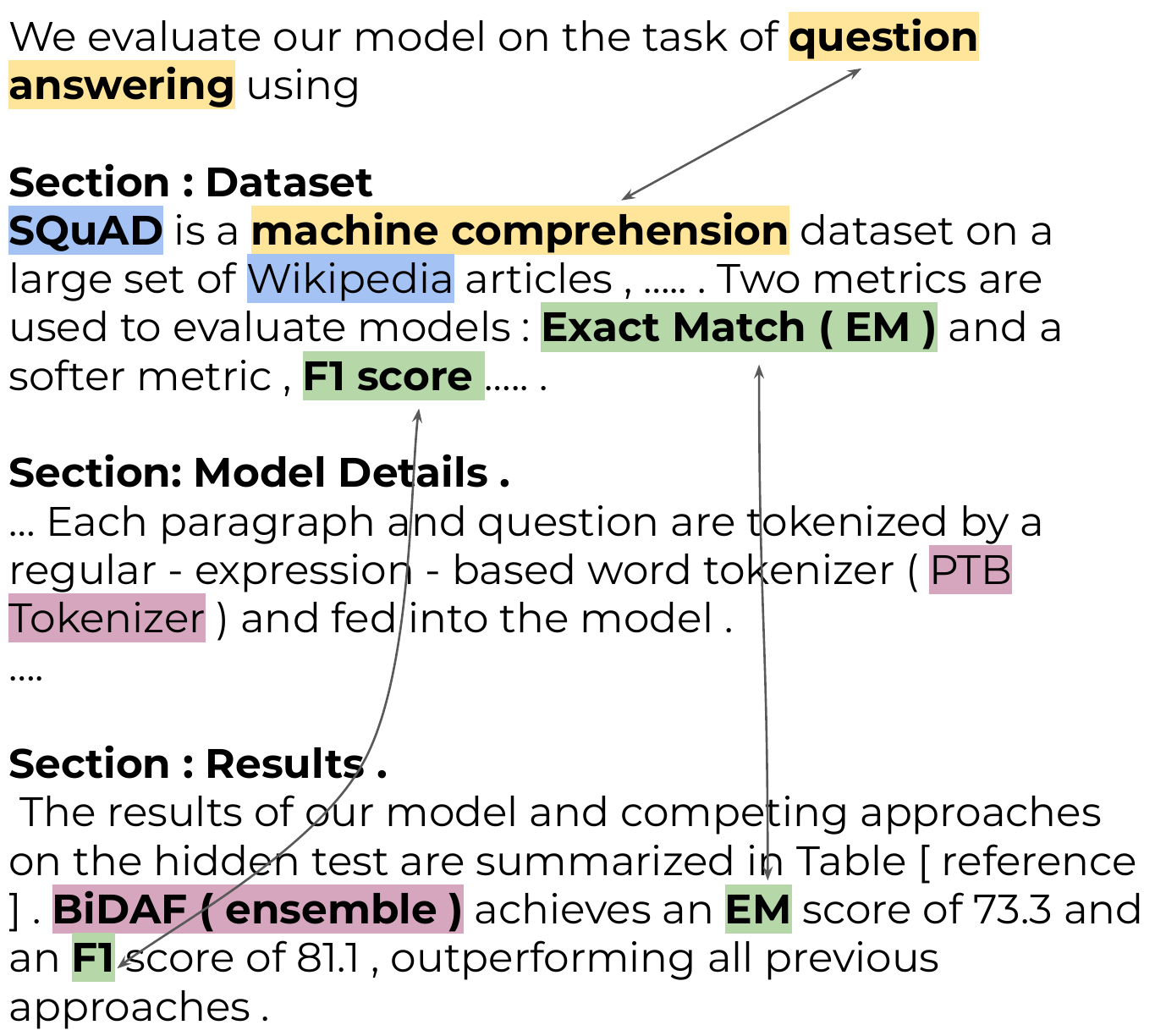}
    \end{boxedminipage}
    \caption{An example showing annotations for entity mentions (\colorbox{dataset}{Dataset}, \colorbox{metric}{Metric}, \colorbox{task}{Task}, \colorbox{method}{Method}), coreferences (indicated by arrows), salient entities (bold), and $N$-ary relation (SQuaD, Machine Comprehension, BiDAF (ensemble), EM/F1) that can only be extracted by aggregating information across sections.}
    \label{fig:main-fig}
\end{figure}

Creating datasets for information extraction at the document level is challenging because it requires domain expertise and considerable annotation effort to comprehensively annotate a full document for multiple IE tasks. 
In addition to local relationships between entities, 
it requires identifying document-level relationships that go beyond sentences and even sections.
Figure~\ref{fig:main-fig} shows an example 
of such document level relation { (Dataset: {\it SQuAD}, Metric: {\it EM}, Method: {\it BiDAF}, Task:{\it machine comprehension})}. 

In this paper, we introduce \dataset, a new comprehensive dataset for information extraction from scientific articles. Our dataset focuses on the task of identifying the main results of a scientific article  as a tuple
({Dataset}, {Metric}, {Task}, {Method}) from raw text. 
It consists of three major subtasks, identifying individual entities, 
their document level relationships,
and predicting their saliency in the document (i.e., entities that take part in the results of the article and are not merely, for example, mentioned in Related Work). 
Our dataset is fully annotated with entities, their mentions, their coreferences, and their document level relations.

To overcome the annotation challenges for large documents, we perform both automatic and manual annotations, leveraging external scientific knowledge bases. An automatic annotation stage identifies candidate mentions of entities with high recall,
then an expert annotator  corrects these extracted mentions by referring to 
the text of the article and an external knowledge base.\footnote{Papers with Code: \url{paperswithcode.com}}
This strategy significantly reduces the time necessary to fully annotate large documents for multiple IE tasks. 

In addition, we introduce a neural model as a strong baseline 
to perform this task end-to-end. 
Our model identifies mentions,  their saliency, and their coreference links. It then clusters salient mentions into entities and identifies document level relations. 
We did not find other models that can perform the full task,
so we evaluated existing state-of-the-art models on 
subtasks, and found our baseline model to outperform them. Experiments also show that 
our end-to-end document level IE task is challenging, with 
the most challenging subtasks being identifying salient entities, and 
to a lesser extent, discovering document level relations.

The contributions of our paper are as follows, 
\begin{enumerate*}
    \item we introduce \dataset, a dataset that evaluates
a comprehensive list of IE tasks, including $N$-ary relations that span long documents.
This is a unique setting compared to prior work that focuses on short paragraphs or a single IE task.
\item We develop a baseline model that, to the best of our knowledge, is the first attempt
toward a neural full document IE. Our analysis emphasizes the need for better
IE models that can overcome the new challenges posed by our dataset. 
\end{enumerate*}
We invite the research community to focus on this important, challenging task.

\section{Related Work}
\label{sec:related_work}
\paragraph{Scientific IE} In recent years, there has been multiple attempts to automatically extract structured information from scientific articles. These types of extractions include
citation analysis~\cite{Jurgens2018MeasuringTE,cohan2019structural},
identifying entities and relations~\cite{augenstein-etal-2017-semeval,Dygie,luanentity}, 
and unsupervised detection of entities and their coreference information~\cite{Tsai2013ConceptbasedAO}.

Most structured extraction tasks from among these have revolved around extraction from sentences or abstracts of the articles. A recent example is \scierc~\cite{luan2018multitask}, 
a dataset of 500 richly annotated scientific abstracts containing mention spans and their types, coreference information between mentions, and binary relations annotations. 
We use \scierc to bootstrap our data annotation procedure (Section~\ref{sec:dataset_constrcut}).

There has been a lack of comprehensive IE datasets annotated at the document level. 
Recent work by~\citet{hou2019identification,jia2019document} tried to rectify this by using distant supervision annotations to build datasets for document-level 
relation extraction. 
In both datasets, the task of relation extraction is formulated as a binary classification to check if a triplet of ground-truth entities is expressed in the document or not.
Instead, our work focuses on a comprehensive list of information extraction tasks ``from scratch'', where the input is the raw document. 
This makes the IE model 
more interesting as it requires to perform entity extraction, coreference resolution, saliency detection
in addition to the relation extraction.\footnote{Another approach is to perform entity extraction then use the binary classification approach with a list of all possible combinations of relation tuples. This might work for short documents, but it is intractable for long documents because of the large number of entities. }

\paragraph{General IE} Most work in general domain IE focus on sentence-level information extraction~\cite{Stanovsky2018NAACL,Qin2018RobustDS,Jie2019DependencyGuidedLF}.
Recently, however, \citet{yao-etal-2019-docred} introduced DocRED, a dataset
of cross-sentence relation extractions on Wikipedia paragraphs. The paragraphs are of a comparable length to that of \scierc, which is significantly shorter than documents in our dataset.

Previous IE work on the TAC KBP competitions~\cite{Ellis2017OverviewOL,getman2018laying} comprise
multiple knowledge base population tasks. Our task can be considered a variant of the TAC KBP ``cold start'' task that discovers new entities and entity attributes (slot filling) from scratch.  Two aspects of our task make it more interesting, 1) our model needs to be able to extract facts that are mentioned once or twice rather than rely on the redundancy of information in their documents (e.g~\citet{Rahman2016TACK2}), 2) TAC KBP relations are usually sentence-level binary relations between a query entity and an attribute  (e.g~\citet{Angeli2015BootstrappedST}), while our relations are 4-ary, span the whole document, and can't be split into multiple binary relations as discussed in Section~\ref{sec:task_def}. 

\paragraph{End-to-End Neural IE models} 
With neural networks, a few end-to-end models have been proposed that perform multiple IE tasks jointly~\cite{miwa-bansal-2016-end,luan2018multitask,wadden2019entity}. 
The closest to our work is \dygie~\cite{wadden2019entity}, which does named entity recognition, binary relation extraction, and event extraction in one model. \dygie is a span-enumeration based model which works well for short paragraphs but does not scale well to long documents. Instead, we use a CRF sequence tagger, which scales well. Our model also extracts 4-ary relations between \emph{salient entity clusters}, which requires a more global view of the document than that needed to extract binary relations between all pairs of entity mentions.

\section{Document-Level IE}
Our goal is to extend sentence-level IE to documents and construct a dataset for document-level information extraction from scientific articles. 
This section defines the IE tasks we address, and describe the details of building our \dataset\ dataset. 
\subsection{Task Definition}
\label{sec:task_def}

\paragraph{Entity Recognition} Our entities are abstract objects of type Method, Task, Metric, or Dataset that appear as text in a scientific article. We define ``mentions'' (or spans) as a specific instantiation of the entity in the text -- this could be the actual name of the entity, its abbreviation, etc. The entity recognition task is to identify ``entity mentions'' and classify them with their types.

\paragraph{Salient Entity Identification} Entities appear in a scientific article are not equally important. For example, a task mentioned in the related work section is less important than the main task of the article.
In our case, salient entity identification refers to finding if
an entity is taking part in the article evaluation.
Salient Datasets, Metrics, Tasks, and Methods are those needed to describe the article's results.
For the rest of this paper, we will use the term \emph{salient} to refer to
entities that belong to a result relation tuple. 

\paragraph{Coreference} is the task of identifying a {\it cluster} of mentions of an entity (or a salient entity) that are coreferred in a single document. 

\paragraph{Relation Extraction} is the task of extracting $N$-ary relations between entities in a scientific article. We are interested in discovering binary, 3-ary, and 4-ary relations between a collection of entities of type (Dataset, Method, Metric, and Task). It is important to note that this 4-ary relation can't be split into multiple binary relations because, e.g., a dataset might have multiple tasks, and each one has its own metric, so the metric cannot be decided solely based on the dataset or the task.

\subsection{Dataset Construction} 
\label{sec:dataset_constrcut}
Document-level information extraction requires a global understanding of the full document to annotate entities, their relations, and their saliency.  However, annotating a scientific article is time-consuming and requires expert annotators. 
This section explains our method for building our \dataset dataset with little annotation effort. It combines distant supervision from an existing KB and noisy automatic labeling, to provide a much simpler annotation task.

\paragraph{Existing KB: Papers with Code}
Papers with Code (PwC)\footnote{\url{https://github.com/paperswithcode/paperswithcode-data}} is a  publicly available corpus of 1,170 articles published in ML conferences annotated with result five-tuples of (Dataset, Metric, Method, Task, Score).
The PwC curators collected this data from public leaderboards, 
previously curated results by other people, manual annotations, 
and from authors submitting results of their work.

This dataset provides us with distant supervision signal for
a task that  requires document-level 
understanding - extracting result tuples. 
The signal is ``distant''~\cite{Riedel2010ModelingRA} because,
while we know that the PwC result tuple exists in the article, we don't know where exactly it is mentioned (PwC does not provide entity spans, 
and PwC entity names may or may not appear exactly in the document).

\paragraph{PDF preprocessing}
PwC provides arXiv IDs for their papers. To extract raw text and section information, we use LaTeXML (\url{https://dlmf.nist.gov/LaTeXML/}) for papers with latex source (all 438 annotated papers), or use Grobid~\cite{GROBID} for papers in PDF format (only 10\% of remaining papers did not have latex source). LaTeXML allowed us to extract clean document text with no figures / tables / equations. We leave it as future work to augment our dataset with these structured fields. To extract tokens and sentences, we use the SpaCy (\url{https://spacy.io/}) library.

\begin{table}[t]
\centering
\begin{tabular}{lrr}
\toprule
    Statistics (avg per doc)   & \dataset & \scierc \\\midrule
    Words & 5,737 & 130\\
    Sections & 22 & 1 \\
    Mentions & 360 & 16\\
    Salient Entities & 8 & --- \\
    Binary Relations & 16 & 9.4 \\
    4-ary Relations & 5 & ---\\ \bottomrule
\end{tabular}
\caption{Comparison of \dataset with next biggest ML Information Extraction dataset \scierc. \dataset\ consists of 438 documents. All dataset statistics are per-document averages.  57\% of binary and 99\% of 4-ary relations occur across sentences. 20\% binary and 55\% 4-ary relations occur across sections. This highlight the need for document level models. }
\label{tab:dataset_stat}
\end{table}

\begin{table}[t]
    \centering
    \small
    \begin{tabular}{lrrrrr}
    \toprule
    {} &  Dataset &  Metric &  Task &  Method &  Deleted \\
    \midrule
    Dataset &    3.55 &    0.01 &  0.07 &    0.16 &  0.03 \\
Metric   &      0.02 &    7.95 &  0.00 &    0.03 &  0.00 \\
Task     &      0.32 &    0.07 & 17.92 &    0.44 &  0.01 \\
Method   &      0.65 &    0.21 &  0.24 &   53.27 &  0.02 \\
Added     &      2.40 &    1.30 &  2.82 &    8.50 &  - \\
    \bottomrule
    \end{tabular}
    \caption{Confusion Matrix for the mention-level corrections (change type, add span, or delete span). Values are average percentages ``per document'' (not per type). For example, cell at intersection of row \textbf{Metric} and column \textbf{Task} contains document-average percentage of span-type change from Metric to Task. The column \textbf{Deleted} represents percent spans that were deleted. The row \textbf{Added} represents percent spans added. Diagonal represent percent spans of each type that are correctly labeled by the automatic labeling and didn't need to change by the human annotator.}
    \label{tab:confusion-ann}
\end{table}

\paragraph{Automatic Labeling}
Given the length of the document is on the order of 5K tokens, we simplify the 
human annotation task by automatically labeling the data with noisy labels, 
then an expert annotator only needs to fix the labeling mistakes.

One possible way to augment the distant supervision provided by PwC is
finding mention spans of PwC entities. Initial experiments 
showed that this did not work well
because it does not provide enough span-level annotations that the model 
can use to learn to recognize mention spans.

To get more dense span-level information,
we want to label salient (corresponding to PwC entities) and also non-salient spans. We train a standard 
BERT+CRF sequence labeling model on the \scierc dataset (described in Section~\ref{sec:related_work}).
We run this model on each of the documents in the PwC corpus, and it provides us with automatic (but noisy) predictions for mention span identification.

The next step is to find mention spans that correspond to PwC entities. 
For each mention predicted by our \scierc-trained model, we compute a Jaccard similarity with each of the PwC entities. Each mention is linked to the entity if the threshold exceeds a certain $\epsilon$. To determine $\epsilon$, two expert annotators manually went through 10 documents to mark identified mentions with entity names, and $\epsilon$ was chosen such that the probability of this assignment is maximized. We use this threshold to determine a mapping for the remaining 1,170 documents. Given that Jaccard-similarity is a coarse measure of similarity, this step favors high recall over precision.

\paragraph{Human Annotation}
Given this noisily labeled data, we ask our annotator to perform necessary corrections to generate high-quality annotations. Annotators are provided with a list of papers-with-code entities that they need to find in the document, making their annotations deliberate (as opposed to not knowing which entities to annotate). Our annotator deleted and modified types of spans for salient entities (belong to PwC result tuple) and non-salient entities, while only adding missed spans for salient ones. Also, if a mention was linked to a wrong PwC entity, then our annotator was also asked to correct it. Full annotation instructions are provided in Appendix~\ref{sec:anno_guide}. 
 
\subsection{Dataset and Annotation Statistics}

\paragraph{Dataset statistics and Cross-section Relations} 
Using the annotation procedure mentioned above, we build a dataset 
of 438 fully annotated documents. 
Table~\ref{tab:dataset_stat} provides 
dataset statistics and shows the proportion of relations in our dataset that requires reasoning across sentence/section. It shows that the majority of the relations, especially 4-ary relations
span multiple sentences or even multiple sections. An example of such cross-section reasoning can be found in Figure~\ref{fig:main-fig}.

\paragraph{Corrections}
Table~\ref{tab:confusion-ann} provides information about the average number of changes made during the human annotation. It shows that 83\% (sum of diagonal) are correct automatic labels, 15\% (sum of bottom row)  are newly added spans, 2\% are type changes, and a negligible percentage is deleted entities (sum of the last column). Also, on average, 12\% (not in the table) of the final mentions in the document had the wrong PwC links and needed to be corrected, with a majority of changes being removing links from Method spans.

\paragraph{Inter-annotator agreement}
We also asked four experts (Ph.D. students in ML/NLP field) to annotate five documents to compute the inter-annotator agreement. For mention classification, we achieve 95\% average cohen-$\kappa$ scores between each pair of experts and our main annotator.

\paragraph{Annotation Speed}
To measure if automatic labeling is making the human annotation faster, we also asked our annotator to perform annotations on five documents without automatic labeling. We compute the difference in time between these two forms of annotation per entity annotated. Note that here, we only ask our annotator to annotate salient mentions. With the automatic labeling, annotation speed is 1.34 sec per entity time vs. 2.48 sec per entity time on documents without automatic labeling (a 1.85x speedup).
We also observe 24\% improvement in recall of salient mentions by including non-salient mentions, further showing the utility of this approach.

%%%%%%%%%%%%%%%%%%%%%%%%%%%%%%%%%%%%%%%%%%%%%%%%%%%%%%%%%%%%%%%%%%%%%%%%%%%%%%%%%%%%%%%%%%%%%%%%%%%%%%%%%%
%%% Model
%%%%%%%%%%%%%%%%%%%%%%%%%%%%%%%%%%%%%%%%%%%%%%%%%%%%%%%%%%%%%%%%%%%%%%%%%%%%%%%%%%%%%%%%%%%%%%%%%%%%%%%%%%

\section{Model}

\begin{figure*}[t]
    \centering
    \includegraphics[width=.67\linewidth]{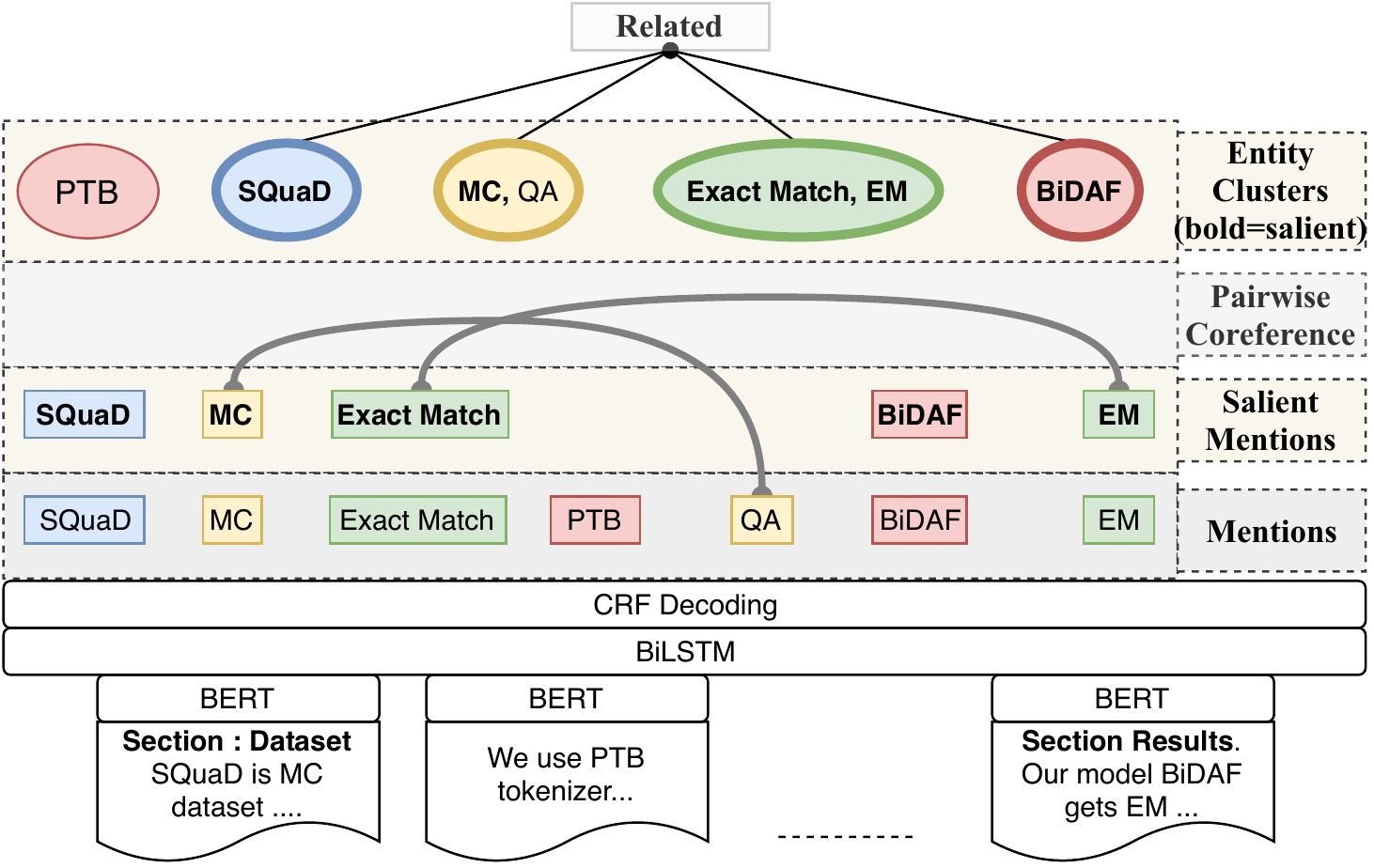}
    \caption{Overview of our model; it uses a two-level BERT+BiLSTM method to get token representations which are passed to a CRF layer to identify mentions. Each mention is classified as being salient or not. A coreference model is trained to cluster these mentions into entities. A final classification layer predicts relationships between  4-tuple of entities (clusters).
    }
    \label{fig:model}
\end{figure*}

We develop a neural model that performs document-level IE tasks jointly in an end-to-end fashion.\footnote{with the exception of coreference resolution}
This section details our model design (also summarized in Figure~\ref{fig:model}).

\vspace{-.1cm}
\paragraph{Document Representation}
An input document $D$ is represented as a list of sections $[s_1, ..., s_{|S|}]$. We encode the document in two steps, section-level, then document-level. We use pretrained contextualized token encodings using SciBERT~\cite{beltagy2019scibert} over each section separately to get embeddings for tokens in that section. \footnote{If the section is bigger than 512 tokens (SciBERT limit), it is broken into 512 token subsections, and each subsection is encoded separately.}
To allow document-level information flow, we concatenate the section-level token embeddings and add a BiLSTM on top of them. This allows the model to take into account cross-section dependencies. Thus for each token $w_i$ in the document, this step outputs an embedding $e_i$.

\paragraph{Mention Identification and Classification}
Given token embeddings, our model applies a sequence tagger that identifies mentions and classifies their types. We train a BIOUL based CRF tagger on top of the BERT-BiLSTM embeddings of words to predict mention spans $m_j$ and their corresponding types. 

\paragraph{Mention Representation} 
Given the words $\{w_{j_1}, ..., w_{j_N}\}$ of a mention $m_j$, 
our model learns a mention embedding $me_j$ of the mention, which will be used in later saliency identification and relation classification steps. The mention embedding is the concatenation of first token embedding $e_{j_1}$, last token embedding $e_{j_N}$ and attention weighted average of all embeddings in the mention span $\sum_{k=1}^{N} \alpha_{j_k} e_{j_k}$, where $e_{j_k}$ is the embedding of word $w_{j_k}$ and $\alpha_{j_k}$ are scalars computed by passing the token embedding through an additive attention layer~\cite{bahdanau2014neural}. We concatenate these embeddings with additional features --- span's relative position in the document, an indicator showing if the sentence containing the mention also contains some marker words like `experiment' or `dataset' and the mention type. 

\paragraph{Salient Mention Classification}
Each mention $m_j$ is classified as being salient or not (i.e., should it belong in a relation tuple) by passing its span embedding $me_j$ through a feedforward layer. Because saliency is a property of entities, not mentions, this mention saliency score is just an input to the salient entity cluster identifications.

\paragraph{Pairwise Coreference Resolution}
The coreference step is given a list of all pairs of identified mentions, and it decides which pair is coreferring. This component is separate from the end-to-end model. It concatenates the ``surface forms'' of two spans $m_i$ and $m_j$, embed them using SciBERT, then use a linear classification layer on top of \texttt{[CLS]} embedding to compute the pairwise coreference score $c_{ij}$. We also tried integrating it into our model, where we classify pairs of ``span embeddings'' (not the surface form) but found the separate model that uses surface forms to work much better. 

\paragraph{Mention clustering}
Given a list of span pairs $m_i$ and $m_j$, and their pairwise coreference scores $c_{ij}$, they are grouped into clusters that can be thought of as representing a single entity. We generate a coreference score matrix for all pairs and perform agglomerative hierarchical clustering~\cite{Inchoate:Ward63} on top of it to get actual clusters. The number of clusters is selected based on the silhouette score~\cite{ROUSSEEUW198753} 
which optimizes for the cohesion and separation of clusters and does not depend on having gold standard cluster labels.
\paragraph{Salient Entity Cluster Identification}
This step filters out clusters from the previous step, and only keep salient clusters for the final relation task. 
To do so, we take a simple approach that identifies a salient cluster as the one in which there is at least one salient mention (as determined previously). The output of this step is a set of clusters $C_1, ..., C_L$ where each cluster $C_i$ is a set of mentions $\{m_{i_1}, ..., m_{i_j}\}$ of the same type.
\paragraph{Relation Extraction} 
Given all the clusters of mentions identified in a document from the previous step, our task now is to determine which of these belong together in a relation. To that end, we follow \cite{jia2019document} methodology.  We consider all candidate binary and 4-tuples of clusters and classify them as expressed or not expressed in the document. Here we describe the classification of 4-ary relations. For binary relation, the method is similar.

Consider such a candidate relation (4-tuple of clusters) $R=(C_1, C_2, C_3, C_4)$ where each $C_i$ is a set of mentions $\{m_{i_1}, ..., m_{i_j}\}$ in the document representing the same entity. We encode this relation into a single vector by following a two-step procedure -- constructing a section embedding and aggregating them to generate a document level embedding. For each section $s$ of the document, we create a section embedding $E_R^s$ for this relation as follows - 

For each cluster $C_i \in R$, we construct its section embedding $E_i^{s}$ by max-pooling span embeddings of the mentions of $C_i$ that occur in section $s$ (along with a learned bias vector $b$ in case no mentions of $C_i$ appear in section $s$). Then the section $s$ embedding of tuple $R$ is $E_R^s = \text{FFN}([E_1^s; E_2^s; E_3^s; E_4^s])$ where $;$ denotes concatenation and FFN is a feedforward network.
 We then construct a document level embedding of $R$, $E_R$ as mean of section embeddings $\frac{1}{|S|}\sum_{s=1}^{|S|} E_R^s$. The final classification for relationship is done by passing the $E_R$ through another FFN, which returns a probability of this tuple expressing a relation in this document.

\paragraph{Training Procedure}
While mention identification, span saliency classification, and relation extraction share the base document and span representation from BERT + BiLSTM and trained jointly, each of these subparts is trained on ground truth input. Note that we require the saliency classification and relation extraction to be independent of mention identification task since the output of this task (essentially the span of mention text) is non-differentiable. \footnote{It is conceivable that mixing the gold mention spans with predicted mention spans might give an improvement in performance; therefore, we leave this as future work. } 
The model jointly optimizes three losses, 
negative log-likelihood for mention identification, 
binary cross-entropy for saliency classification,
and binary cross-entropy for relation extraction,
with all three losses weighted equally.

\section{Evaluation}
We compare our model with other recently introduced models. Since we cannot apply previous models directly to our task, we evaluate on subtasks of our 
dataset 
and also evaluate on \scierc (Section~\ref{sec:baseline}). 
The other goal of the evaluation is to establish a baseline performance on our dataset and to provide insights into the difficulty of each subtask. To that end, we evaluate the performance of each component separately (Section~\ref{sec:components}), and in the overall end-to-end system (Section~\ref{sec:endtoend}). In addition, we perform diagnostic experiments to identify the bottlenecks in the model performance. 
We report experimental setup and hyperparameters in appendix~\ref{sec:train}.

\subsection{Evaluation Metrics}

\noindent
\textbf{Mention Identification} is a sequence labeling task, which we evaluate using the standard macro average F1 score of exact matches of all mention types.

\noindent \textbf{Salient Mentions} and  \noindent \textbf{Pairwise Coreference} are binary classification tasks which we evaluate using the F1 score. 

\noindent
\textbf{Salient Entity Clustering} evaluation relies on some mapping between the set of predicted clusters and gold clusters. 
Given a predicted cluster $\mathcal{P}$ and a gold cluster $\mathcal{G}$, we consider $\mathcal{P}$ to match $\mathcal{G}$ if more than 50\% of $\mathcal{P}$'s mentions belong to $\mathcal{G}$,\footnote{We consider two mention spans to be a match if their Jaccard similarity is greater than 0.5.}  that is $\frac{|\mathcal{P} \cap \mathcal{G}|}{|\mathcal{P}|} > 0.5$. The 0.5 threshold enjoys the property that, assuming all predicted clusters are disjoint from each other (which is the case by construction) and gold clusters are disjoint from each other (which is the case for 98.5\% of them), a single predicted cluster can be assigned to \emph{atmost one} gold cluster. This maps the set of predicted clusters to gold clusters, and given the mapping, it is straightforward to use the F1 score to evaluate predictions. This procedure optimizes for identifying all gold clusters even if they are broken into multiple predicted clusters. 

\noindent
\textbf{Relation Extraction} evaluation relies on the same mapping used in the evaluation of salient entity clustering. Under such mapping, each predicted $N$-ary relation can be compared with gold relations, and decide if they match or not. This becomes a binary classification task that we evaluate with positive class F1 score. 
We report F1 scores for binary and 4-ary relation tuples. We get binary relations by splitting each 4-ary relation into six binary ones. 

\subsection{Comparing with Baselines}\label{sec:baseline}
We compare our model with \textsc{DyGIE++}~\cite{wadden2019entity} and DocTAET~\cite{hou2019identification} on subtasks of our \dataset dataset and on the \scierc dataset wherever they apply. Our results show that only our model can perform all the subtasks in an end-to-end fashion and performs better than or on par with these baselines on respective subtasks.

\subsubsection{Evaluation on \dataset}
\paragraph{\textsc{DyGIE++}}~\cite{wadden2019entity} is an end-to-end model for entity and binary relation extraction (check Section~\ref{sec:related_work} for details). 
Being a span enumeration type model, \textsc{DyGIE++} only works on paragraph level texts and extracts relations between mentions in the same sentence only. Therefore, we subdivide \dataset documents into sections and formulate each section as a single training example. We assume all entities in relations returned by \textsc{DyGIE++} are salient. We map each binary mention-level relation returned to entity-level by mapping the span to its gold cluster label if it appears in one.  
We consider 3 training configurations of \textsc{DyGIE++}, 
\begin{enumerate*}
\item trained only on the abstracts in our dataset,
\item trained on all sections of the documents in our dataset.
\item trained on \scierc dataset (still evaluated on our dataset),
\end{enumerate*}
At test time, we evaluate the model on all sections of the documents in the test set.

Results in Table~\ref{tab:sota} show that we perform generally better than \textsc{DyGIE++}. The performance on end-to-end binary relations shows the utility of incorporating a document level model for cross-section relations, rather than predicting on individual sections. Specifically, We observe a large difference in recall, which agrees with the fact that 55\% of binary relation occur across sentence level. 
\textsc{DyGIE++} (All sections) were not able to identify any binary relations 
because 80\% of training examples have no sentence level binary relations, pushing 
the model towards predicting very few relations.
In contrast, training on \scierc (and evaluating on \dataset) gives better results
because it is still able to find the few sentence-level relations. 

\paragraph{DocTAET}~\cite{hou2019identification} is a document-level 
relation classification model that is given a document and a relation tuple to classify if it is expressed in the document. It is formulated as an entailment task with the information encoded as \texttt{[CLS] document [SEP] relation} in a BERT style model. This is equivalent to the last step of our model but with gold salient entity clusters as input. 
Table~\ref{tab:sota} shows the result on this subtask, 
and it shows that our relation model gives comparable performance (in terms of positive class F1 score) to that of DocTAET.

\begin{table}[t]
    \centering
    \small
    \begin{tabular}{l|ccc}
    \toprule
      Model & P & R & F1 \\
      \midrule
      \multicolumn{4}{c}{Mention Identification} \\ \hline
      \textsc{DyGIE++} & 0.703 & 0.676 & 0.678 \\
      Our Model & 0.707 & 0.717 & \textbf{0.712} \\ \hline
       \multicolumn{4}{c}{End-to-end binary relations} \\ \hline
       \textsc{DyGIE++} (Abstracts Only) & 0.003 & 0.001 & 0.002 \\ 
       \textsc{DyGIE++} (All sections)& 0.000& 0.000 & 0.000  \\
       \textsc{DyGIE++} (\scierc)& 0.029 & 0.128 & 0.038 \\ 
       Our Model  & \textbf{0.065} & \textbf{0.411} & \textbf{0.096} \\ \hline
      \multicolumn{4}{c}{4-ary relation extraction only} \\ \hline
        DocTAET  & 0.477 & \textbf{0.885} & 0.619\\
        Our Model & \textbf{0.531} & 0.718 & 0.611 \\ 
      \bottomrule 
    \end{tabular}
    \caption{Evaluating state-of-the-art models on subtasks of \dataset dataset 
    because we did not find an existing model that can perform the end-to-end task.}
    \label{tab:sota}
\end{table}

\subsubsection{Evaluation on \scierc}

\begin{table}[]
    \centering
    \small
    \begin{tabular}{l|l|ccc}
    \toprule
       Task & Model &  P & R & F1\\
        \midrule
       Mention Ident. & \dygie &  0.676 & 0.694& 0.685  \\
        & Our Model & 0.637 & 0.640 &  0.638\\
        \midrule
       Pairwise Coref. & \dygie & 0.577 & 0.455 & 0.476 \\
       and Clustering & Our Model & 0.187 & 0.552 & 0.255 \\
        \bottomrule
    \end{tabular}
    \caption{Comparison of \dygie with our model on various subtasks of \scierc dataset}
    \label{tab:scierc}
\end{table}

Table~\ref{tab:scierc} summarizes the results of evaluating our model and 
 \textsc{DyGIE++} on the \scierc dataset.
For mention identification, our model performance is a bit worse mostly 
because \scierc has overlapping entities that a CRF-based model like ours can not 
handle. 
For the task of identifying coreference clusters, we perform significantly worse than \textsc{DyGIE++}'s end-to-end model. This provides future avenues towards improving coreference resolution for \dataset by incorporating it in an end-to-end fashion. 

\subsection{Component-wise Evaluation}
\label{sec:components}
\begin{table}[t]
    \centering
    \small
    \begin{tabular}{l|ccc}
    \toprule
      Task & P & R & F1 \\
      \midrule
      \multicolumn{4}{c}{Component-wise (gold Input)} \\ \hline
      Mention Identification   & 0.707 & 0.717 & 0.712  \\
      Pairwise Coreference   & 0.861 & 0.852 & 0.856 \\
      Salient Mentions  & 0.575 & 0.584 & 0.579 \\
      Salient Entity Clusters   & 1.000 & 0.984 & 0.987 \\
      Binary Relations  & 0.820 & 0.440 & 0.570 \\
      4-ary Relations  & 0.531 & 0.718 & 0.611 \\ \hline
    \multicolumn{4}{c}{End-to-end (predicted input)} \\ \hline
      Salient Entity Clusters &  0.223 & 0.600 & 0.307  \\
      Binary Relations  & 0.065 & 0.411 & 0.096 \\
      4-ary Relations  & 0.007 & 0.173 & 0.008  \\  \hline
          \multicolumn{4}{c}{End-to-end (gold salient clustering)} \\ \hline
      Salient Entity Clusters & 0.776 & 0.614 & 0.668  \\
      Binary Relations &  0.372 &    0.328 & 0.334    \\
      4-ary Relations     & 0.310 & 0.281 & 0.268  \\ 
      \bottomrule 
    \end{tabular}
    \caption{Analysis of performance of our model and its subtasks under different evaluation configurations.}
    \label{tab:all_results}
\end{table}

The main contribution of our model is to connect multiple components to perform our end-to-end task. This section evaluates each step of our model separately from all other components. To do so, we feed each component with gold inputs and evaluate the output. This gives us a good picture of the performance of each component without the accumulation of errors. 

The first block of Table~\ref{tab:all_results} summarizes the results of this evaluation setting. We know from Tables~\ref{tab:sota},~\ref{tab:scierc} that 
our mention identification and relation identification components are working well.  
For pairwise coreference resolution, we know from Table~\ref{tab:scierc}
that it needs to be improved, but it is performing well on our dataset likely because the majority of coreferences in our dataset can be performed using only the surface form of the mentions (for example, abbreviation reference).
The worst performing component is identifying salient mentions, which requires information to be aggregated from across the document, something the current neural models lack.\footnote{Performance of Salient Entity Clusters is close to 1.0 because it is a deterministic algorithm (clustering followed by filtering) that gives perfect output given gold input. The reason the recall is not 1.0 as well is because of small inconsistencies in the gold annotations (two distinct entities merged into one).}

\subsection{End-to-End Evaluation}
\label{sec:endtoend}

\paragraph{Evaluation with Predicted Input.} The second block in Table~\ref{tab:all_results} gives results for the end-to-end performance of our model in predicting salient entity clusters, binary relations, and 4-ary relations. 
We noticed that there is quite a drop in the end-to-end performance compared to the component-wise performance. This is particularly clear with relations; even though the relation extraction component performance is reasonably good in isolation, its end-to-end performance is quite low because of the accumulation of errors in previous steps.

\paragraph{Evaluation with Gold Salient Clustering.} ~\\
Through manual error analysis, we found that the identification of salient clusters
is the most problematic step in our model. The third block in Table~\ref{tab:all_results}
quantifies this. In this setting, we run our end-to-end model but with
``gold cluster saliency'' information.
In particular, we predict clusters of mentions using our model (mention identification, pairwise coreference, and mention clustering). Then instead of filtering clusters using our mention saliency score, we keep only those clusters that have any overlap with at least one gold cluster. Predicted clusters that match the same gold cluster are then combined. Finally, we feed those to the relation extraction step of our model. 
Under this setting, we found that the performance of 4-ary relations improves considerably
by more than 10x. This confirms our hypothesis that identifying salient clusters is the key bottleneck in the end-to-end system performance. This is also consistent with the component-wise results that show low performance for salient mentions identification. 

\paragraph{Error Analysis for Identifying Salient Clusters.}
Our error analysis shows that the average number of mentions in a salient cluster classified correctly is 15 mentions, whereas for the misclassified ones is six mentions. This indicates that our model judges the saliency of an entity strongly based on how frequently it is mentioned in the document. While this is a perfectly reasonable signal to rely on, the model seems to trust it more than
the context of the entity mention. For example, in the following snippet,
``\emph{... For each model, we report the test perplexity, the computational budget, the parameter counts, the value of DropProb, and the computational efficiency ....}'', the entity ``\emph{the parameter counts}'' is misclassified as non-salient, as it only appears twice in the document. One possible way to address this issue with salient entity identification is to replace its simple filtering step with a trained model that can do a better job at aggregating evidence from multiple mentions. 

Overall, these results indicate that identifying the saliency of entities in a scientific document is a challenging task. It requires careful document-level analysis, and getting it right is crucial for the performance of an end-to-end document-level IE model.  Also, the difference between results in the third block of the results and the component-wise results indicate that the whole model can benefit from incremental improvements to each component.

\section{Conclusion}
We introduce \dataset, a comprehensive and challenging dataset for information extraction on full documents. We also develop a baseline model for our dataset, which, to the best of our knowledge, is the first attempt toward a neural document level IE that can perform all the necessary subtasks in an end-to-end manner. We show that using a document level model gave a significant improvement in terms of recall, compared to existing paragraph-level approaches. 

This task poses multiple technical and modeling challenges, including
\begin{enumerate*}
    \item the use of transformer-based models on long documents and related device memory issues,
    \item aggregating coreference information from across documents in an end-to-end manner,
    \item identifying salient entities in a document and
    \item performing N-ary relation extraction of these entities.
\end{enumerate*}
Each of these tasks challenges existing methodologies in the information extraction domain, which, by and large, focus on short text sequences. An analysis of the performance of our model emphasizes the need for better document-level models that can overcome the new challenges posed by our dataset. As our research community moves towards document level IE and discourse modeling, we position this dataset as a testing ground to focus on this important and challenging task.

\section*{Acknowledgments}
This research was supported by the ONR MURI
N00014-18-1-2670, ONR N00014-18-1-2826,
DARPA N66001-19-2-4031, and Allen Distinguished Investigator Award. We thank the Semantic Scholar team at AI2, UW NLP, and anonymous reviewers for their insightful comments. We are especially grateful to Kyle Lo for help with Grobid parser, the complete Papers With Code team for making their data publicly available, Dan Weld and Robert Stojnic for helpful discussion and feedback.

\bibliography{anthology,acl2020}
\bibliographystyle{acl_natbib}

\appendix

\section{Model Details}
\label{sec:train}
We divide our 438 annotated documents into training (70\%), validation (30\%) and test set (30\%).
The base document representation of our model is formed by SciBERT-base \cite{beltagy2019scibert} and BiLSTM with 128-d hidden state. We use a dropout of 0.2 after BiLSTM embeddings. All feedforward networks are composed of two hidden layers, each of dimension 128 with gelu activation and with a dropout of 0.2 between layers. For additive attention layer in span representation, we collapse the token embeddings to scalars by passing through the feedforward layer with 128-d hidden state and performing a softmax. We train our model for 30 epochs using Adam optimizer with 1e-3 as learning rate for all non BERT weights and 2e-5 for BERT weights. We use early stopping with a patience value of 7 on the validation set using relation extraction F1 score. All our models were trained using 48Gb Quadro RTX 8000 GPUs. The multitask model takes approximately 3 hrs to train.

For the BERT coreference model, we use SciBERT-base embeddings with two mentions encoded as [CLS] mention 1 [SEP] mention 2 [SEP]. We use a linear layer on top of [CLS] token embedding to compute the mention pair's coreference score.

All our models were implemented in AllenNLP library\cite{Gardner2017AllenNLP}.

\section{Annotation Guidelines}
\label{sec:anno_guide}

Our Annotation guidelines can be found at \url{https://github.com/allenai/SciREX/blob/master/Annotation\%20Guidelines.pdf} Note, for Method type entities, we specifically ask our annotator to break down complex entities into simpler ones before looking for mentions in the text. For example, a method entity \texttt{DLDL+VGG-Face} is composite and broken into two parts \texttt{DLDL} and \texttt{VGG-Face}. Currently, our model considers all mentions of subentities as mentions of the corresponding Method entity. We leave the task of extracting relation between subentities explicitly as future work.

\end{document}